\documentclass[conference]{IEEEtran}
\usepackage{cite}
\usepackage{amsmath,amssymb,amsfonts}
\usepackage{graphicx}
\usepackage{textcomp}

\usepackage[linesnumbered,ruled]{algorithm2e}
\usepackage[noend]{algpseudocode}
\usepackage{eqparbox}

\SetAlgorithmName{Algorithm}{Algorithm}{Algorithm}
\IncMargin{0.5em}
\SetCommentSty{textnormal}
\SetNlSty{}{}{:}
\SetAlgoNlRelativeSize{0}
\SetKwInput{KwGlobal}{Global}
\SetKwInput{KwPrecondition}{Precondition}
\SetKwProg{Proc}{Procedure}{:}{}
\SetKwProg{Func}{Function}{:}{}
\SetKw{And}{and}
\SetKw{Or}{or}
\SetKw{To}{to}
\SetKw{DownTo}{downto}
\SetKw{Break}{break}
\SetKw{Continue}{continue}
\SetKw{SuchThat}{\textit{s.t.}}
\SetKw{WithRespectTo}{\textit{wrt}}
\SetKw{Iff}{\textit{iff.}}
\SetKw{MaxOf}{\textit{max of}}
\SetKw{MinOf}{\textit{min of}}
\SetKwBlock{Match}{match}{}{}

\usepackage{subcaption}
\usepackage{multirow}
\usepackage{adjustbox}
\usepackage{nccmath}
\usepackage{mathtools}
\usepackage{amsthm}
\usepackage{lipsum}
\usepackage{xcolor}
\DeclareMathOperator{\E}{\mathbb{E}}
\newcommand{\softmax}{\mathrm{softmax}}
\DeclareMathOperator*{\argmax}{arg\,max}
\DeclareMathOperator*{\argmin}{arg\,min}
\theoremstyle{definition}

\usepackage[capitalize,noabbrev]{cleveref}

\def\BibTeX{{\rm B\kern-.05em{\sc i\kern-.025em b}\kern-.08em
    T\kern-.1667em\lower.7ex\hbox{E}\kern-.125emX}}
\begin{document}

\title{DSAC-C: Robust Discrete Soft-Actor Critic \\ for Distribution Shift Scenarios\\
\thanks{Identify applicable funding agency here. If none, delete this.}
}

\author{\IEEEauthorblockN{Dexter Neo, Tsuhan Chen}
\IEEEauthorblockA{\textit{School of Computing} \\
\textit{National University of Singapore}\\
Singapore \\
\{e0534450@u.nus.edu, tsuhan@nus.edu.sg\}}
}

\maketitle

\begin{abstract}
We present a novel extension to the family of Soft Actor-Critic (SAC) algorithms. We argue that based on the Maximum Entropy Principle, discrete SAC can be further improved via additional statistical constraints derived from a surrogate critic policy. Furthermore, by perturbing these constraints we demonstrate that an added robustness can be provided against potentially unseen or adversarial environments, which are essential for safe deployment of reinforcement learning agents in the real-world. We provide theoretical analysis and show empirical results on low data regimes for both in-distribution and out-of-distribution variants of Atari 2600 games. 
\end{abstract}

\begin{IEEEkeywords}
reinforcement learning, robustness, soft actor-critic
\end{IEEEkeywords}

\section{Introduction}
Model-free deep reinforcement learning (RL) algorithms have received much success in a large number of challenging applications such as video games \cite{mnih2013playing, Silver2016MasteringTG} and robotic control tasks \cite{Schulman2015TrustRP, Gu2016DeepRL}. The usage of neural networks as function-approximators in deep RL has given rise to promising potential in automating control and decision making tasks. 

Unfortunately, current RL algorithms often make the closed-world assumption, i.e., state-transitions that follow a Markov Decision Process which is aligned in-distribution (ID) to the training experience. Which can cause RL agents to fail during deployment, since they do not have the ability to identify or preserve the learnt policy when encountering unseen out-of-distribution (OOD) states. These unexpected OOD states can lead to potentially dangerous situations. For e.g., a self-driving agent learned on a clean urban environment, may potentially fail catastrophically when deployed in a mountainous region. Furthermore, RL agents are often expensive to train \cite{obando2020revisiting} and we ideally want our agents to remain robust despite the potential unknown changes in environments.

\begin{figure}[!tb]
\centering
\small
\includegraphics[width=\columnwidth]{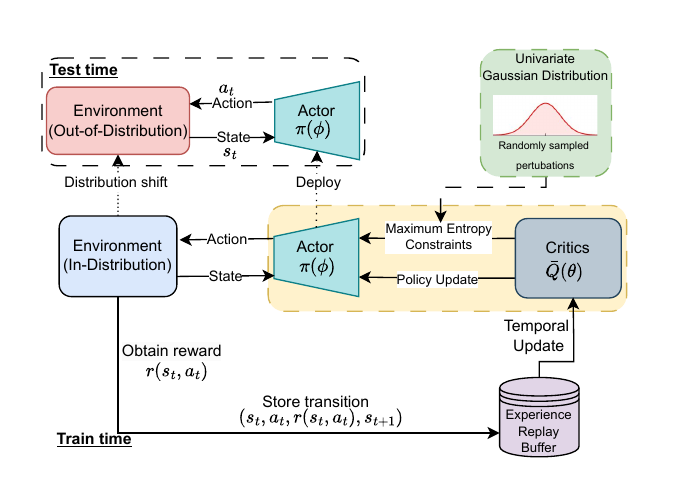}
\caption{Using maximum entropy constraints, DSAC achieves higher performance - even on OOD scenarios.}
\label{fig:fig1}
\end{figure}

\begin{figure*}[!tb]
\small
\centering
\includegraphics[width=\textwidth]{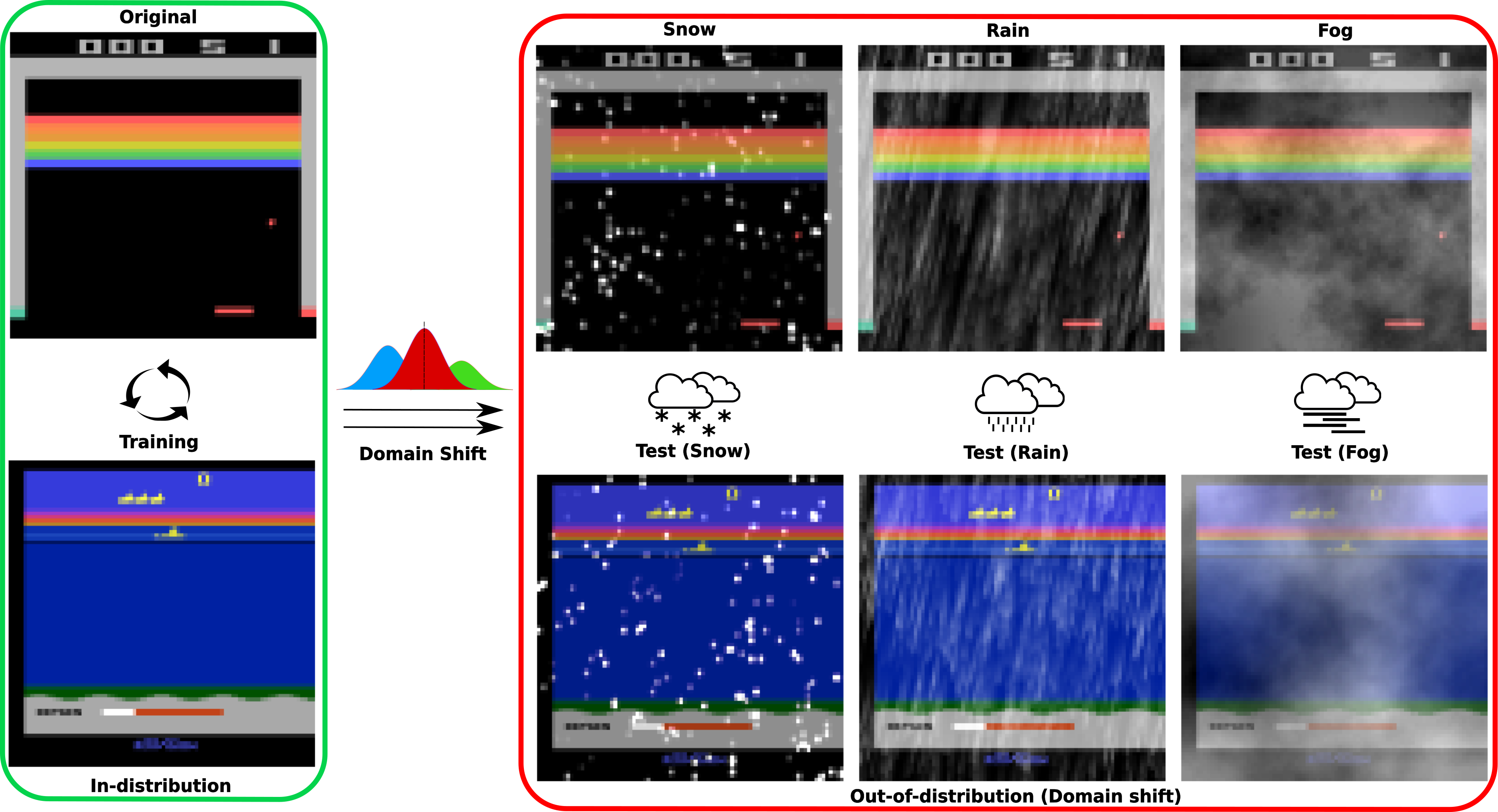}
\caption{Screenshots of two Atari games, Breakout (top) and Seaquest (bottom). Agents are learned using in-distribution environments but evaluated on naturally occurring out-of-distributions shifts, emulating possible scenarios during deployment.}
\label{fig:noisy_atari}
\end{figure*}

Recent solutions to modern deep RL tasks can be broadly organized into three categories of algorithms.
\begin{enumerate}
    \item \textbf{Policy-gradient:} TRPO, \cite{Schulman2015TrustRP}, PPO \cite{Schulman2017ProximalPO}. These require new samples to be collected in order to compute the gradients for policy update. Which may lead to poor sample efficiency, with the number of gradient steps needed to learn an optimal policy tied to the task complexity.
    
    \item \textbf{Value-based:} DQN \cite{mnih2013playing}, Double DQN \cite{Hasselt2015DeepRL}, C51 \cite{Bellemare2017ADP}, dueling networks \cite{Wang2015DuelingNA} and Rainbow \cite{Hessel2017RainbowCI}. These aim to provide value estimates of a given state and are considered off-policy since the use of experience replay is involved. The combination of neural networks and off-policy learning can be problematic for stability and convergence \cite{Maei2009ConvergentTL}. It is worth mentioning that Rainbow is a combination of the aforementioned value-based algorithms and prioritized experience replay \cite{Schaul2015PrioritizedER}, multi-step bootstrap targets \cite{Sutton1998} and exploration with parameter noise \cite{Fortunato2017NoisyNF}.
    
    \item \textbf{Actor-Critic:} These seek to combine both policy and value based methods A3C \cite{Mnih2016AsynchronousMF}, IMPALA \cite{Espeholt2018IMPALASD}, DDPG \cite{Lillicrap2015ContinuousCW}, TD3 \cite{Fujimoto2018AddressingFA} but can be brittle and sensitive to hyperparameter settings.
\end{enumerate}

Although, these algorithms have substantially progressed solving deep RL tasks. The study of OOD RL tasks remains relatively unexplored with many previous works focusing on value-based methods \cite{NEURIPS2021_dbb42293}. In contrast, we ask the question on how would soft actor-critic methods perform on OOD discrete tasks?

Originally proposed for continuous RL settings, SAC \cite{Haarnoja2018SoftAC} was designed to tackle the many underlying issues of TD3 and DDPG. Based on the maximum entropy framework, the authors propose to append an additional entropy term to the objective function. The entropy term naturally improves exploration and has achieved state-of-the-art performance on continuous RL benchmarks such as Mujoco \cite{todorov2012mujoco}. Although SAC's performance on continuous action spaces has been largely successful, its discrete variant has not received much attention. Furthermore, many tasks involve discrete actions and require an alternate version of SAC. 

The translation of SAC from continuous to discrete actions was first proposed by \cite{Christodoulou2019SoftAF} where a direct discretization of the action spaces is applied. Many authors believe that the maximum entropy SAC framework can benefit both continuous and discrete domains. However, empirical experiments from \cite{Xu2021TargetEA, Zhou2022RevisitingDS} suggests that discrete SAC (DSAC) suffers from instability issues and underperforms on discrete benchmarks such as Atari ALE 2600 \cite{Bellemare2012TheAL, Sedlmeier2019UncertaintyBasedOD}. 

Contrary to previous works, we revisit the classical idea of the Maximum Entropy framework \cite{MaxEnt} and demonstrate how additional constraints can be used to improve DSAC's performance. 
We provide systematic comparisons on agent performance and behaviours on a suite of ID and makeshift OOD Atari environments. \cref{fig:noisy_atari} shows the problem setup for potential OOD states that agents could face during deployment, such as "Snow", "Rain" or "Fog" mimicking issues agents could face in the open-world. Our contributions are summarized as follows:

\begin{enumerate}
    \item \textbf{Constrained Maximum Entropy:} We revisit a theoretical link between the Principle of Maximum Entropy and DSAC, introducing constraints that can improve agent performance.
    \item \textbf{Expected Value Targets:} We introduce a simple surrogate policy and perturbed expected value targets that are derived directly from the soft critic. 
    \item \textbf{DSAC-C Lagrangian:} We propose the mean and variance constraint forms of the discrete Soft Actor-Critic algorithm complete with automated Lagrange multiplier tuning for off-policy, model-free deep reinforcement learning.
    \item \textbf{Evaluation on OOD Atari 2600 suite:} Our experiments show that our approach can be used to improve SAC on both ID and OOD discrete RL benchmarks.
\end{enumerate}
\section{Related Work}

\textbf{Discrete Soft-Actor Critic} 
Vanilla discrete SAC \cite{Christodoulou2019SoftAF} was first introduced by directly adapting the action domain from continuous to discrete. However, discrete SAC has largely been unsuccessful and many authors have hypothesized different reasons to its poor performance. For e.g., \cite{Xu2021TargetEA} proposed a scheduling method to improve DSAC's target entropy and temperature tuning. While \cite{Zhou2022RevisitingDS} propose entropy penalty and average Q-clipping to stabilize the policy and Q-value under-estimation.

\noindent \textbf{Offline OOD RL}
Previous works on OOD RL often look at value-based methods, where issues stem from the assignment of high Q-value estimates for OOD states \cite{Sedlmeier2019UncertaintyBasedOD}. Strategies imposed aim to tackle the variances in Q-values such as model ensembles \cite{osband_2018}, placing lower bounds and constraining action selection \cite{kumar_2019,kumar_2020} during Q-learning. In this paper, we focus on DSAC and its robustness towards distribution shifts of the state space during deployment.


\noindent \textbf{Maximum Entropy} In supervised learning, the MaxEnt method provides additional benefits such as improving regularization and model calibration \cite{mukhoti2020calibrating, neo2024maxent}. For reinforcement learning, the MaxEnt method has provided benefits such as robustness against perturbations in environments and reward signal \cite{eysenbach2022maximum} and soft policy gradients \cite{Shi2019SoftPG}.

\section{Preliminaries}
\subsection{Notation}
We provide an overview of the notations and definitions used for infinite-horizon Markov decision processes (MDP). For a given standard MDP defined by the tuple ($\mathcal{S}, \mathcal{A}, \mathcal{T}, r, \gamma$), the state-action space is represented by $\mathcal{S}, \mathcal{A}$, with $\mathcal{T}: S \times S \times A \rightarrow [0, \infty)$ given as the transition probability from the current state $s_t \in \mathcal{S}$ to the next state $s_{t+1} \in \mathcal{S}$ for action $a_t \in \mathcal{A}$ at timestep $t$. For each transition, the environment emits an immediate reward signal $r(s_t, a_t)$ and $\gamma \in [0,1]$ is the discount factor. For deep actor-critic algorithms, the actor network $\phi$ learns a stochastic policy $\pi_\phi(a_t|s_t)$ for a given state from the gradients provided by the Q-value estimates $Q_\theta(s_t,a_t)$ of the critic network $\theta$. In essence, the actor behaves as an approximate sampler, which attempts to estimate the true posterior of the given MDP, whilst the critic attempts to provide the best values estimates for a given state-action pair.
\subsection{Principle of Maximum Entropy}
The Principle of Maximum Entropy\footnote{~We refer readers to \cite{MaxEnt} for further reading regarding the Principle of Maximum Entropy.} (MaxEnt) is a probability distribution with the maximum entropy subject to the expected characteristic constraints. Formally, the general form of the MaxEnt method is given by the following objective:
\label{max_ent_method}
\begin{equation}
\begin{split}
&\max \mathbb{H}(p(y_n|x_n)) =  \E{[ -\log p(y_n|x_n) ]} \\
&\text{s.t. } \E{[f_i(y)]} \le c_i \text{ for all constraints} f_i(y)
\end{split}
\end{equation}
where $p(y_n|x_n)$ are the supervised learning estimates of the posterior distribution for the $n$th sample in the minibatch and $f_i(y)$ is an arbitrary characteristic function of the possible class outcomes $y_n$ given input $x_n$ for each respective constraint $c_i$. The probability weighted characteristic functions are in the form of $\E_{y\sim p}{[f_i(y)]} = \sum_y p(y) f_i(y)$. 

\subsection{Soft-Actor Critic}
Contrary to standard RL problems, where the goal is to maximize the expected sum of rewards $\E_{(s_t, a_t)}[\gamma^t r(s_t, a_t)]$. The goal in maximum entropy reinforcement learning is to maximize the expected sum of rewards along with an additional entropy term. The augmented maximum entropy objective favours stochastic policies \cite{Haarnoja2018SoftAC} with its general definition given by the following expectation \cite{Ziebart2010ModelingPA}:
\begin{equation}
    \pi^* = \argmax_\pi \sum^T_{t=0} \E_{(s_t, a_t)}[\gamma^t r(s_t, a_t) + \alpha \mathbb{H}(\pi(.|s_t))]
\label{max_ent_rl objective}
\end{equation}
where $\pi^*$ denotes the optimal policy for the MDP and the entropy function for the policy given as $\mathbb{H} = -\alpha \sum_\mathcal{A} \pi(.|s_t)\log \pi(.|s_t)$.  The temperature hyperparameter $\alpha$ is used to control the strength of the entropy term. In order to optimize the learning objective in \cref{max_ent_rl objective}, actor-critic algorithms are modified to incorporate the Maximum Entropy Method, alternating between 1.) Policy evaluation; where the critic is updated according to a soft Bellman target and 2.) Policy improvement; where the actor is updated via minimizing the Kullback-Leibler (KL) divergence between $\pi_\phi(a_t|s_t)$ and $Q_\theta(s_{t}, a_{t})$. 3.) Automatic tuning of $\alpha$; where the temperature parameter is updated according to constraints placed.

\noindent \textbf{Policy Evaluation:} For a given policy and state, the probability weighted value estimates for the SAC algorithm are: 
\begin{equation}
     V(s_t) = \E_{a_t \sim \pi} [\min_{i=1,2} Q_{\theta_i(s_{t+1}, a_{t+1})} - \alpha \log(\pi_\phi(s_t)) ]
\label{soft_value_estimate}
\end{equation}
where $V(s_t)$ is evaluated using two critics $\theta_i$ by taking the $\min$ value of the double-critic trick. The critics are learned by computing the tabular targets using the Bellman update equation:
\begin{equation}
\begin{split}
    & J_Q(\theta) = \E_{(s_t, a_t) \sim D} [(Q_\theta(s_t, a_t) \\
    & -(r(s_t, a_t) + \gamma \E_{s_{t+1} \sim \mathcal{T}} V(s_{t+1}) )^2]
\end{split}
\end{equation}
with the critic update given by the method of \textit{temporal differences} \cite{Sutton1998} sampled from the experiences collected and stored in the replay buffer $D$.

\noindent \textbf{Policy Improvement:} For the policy improvement step, the actor is updated in the direction that maximizes the rewards learned by the critic. The policy improvement step is given by minimizing the KL divergence between the actor and critic:
\begin{equation}
     \pi_{new} = \argmin D_{KL} [\pi(.|s_t) || \frac{ \exp (\frac{1}{\alpha} Q^\pi_{old}(s_t) )  }{Z^\pi_{old}(s_t)} ]
\end{equation}
For continuous action tasks, the policy is modified using the reparameterization trick, such that the actor outputs a mean and standard deviation which are then used to define a Gaussian policy. In contrast for discrete tasks, the actor is required to output an action probability distribution $\pi_\theta(a_t|s_t)$ instead of a probability density. Therefore, there is no need for the reparameterization trick or the policy to be in the form of a mean and variance. We can instead directly estimate the action distribution by applying the $\softmax$ function such that $\sum_A \pi_\phi(a_t|s_t) = 1$. The parameters of the actor are learned by minimizing the approximated form of the expected KL-divergence with the temperature parameter $\alpha$:
\begin{equation}
    J_\pi(\phi) = \E_{s_t \sim D} [\alpha\log( \pi_\phi(s_t) ) - Q_\theta(s_t, a_t) ]
\label{cont_actor_update}
\end{equation}
\textbf{Automatic Entropy Adjustment:}
The temperature parameter $\alpha$ regulates the actor's entropy during exploration. Enforcing the entropy term to a fixed temperature is regarded as a poor solution since the actor should be free to explore more across uncertain states and behave deterministic in states which are certain. \cite{Haarnoja2018SoftAA} proposed the following Lagrangian optimization problem for obtaining $\alpha$:
\begin{equation}
\begin{split}
    &\max \E = [ \sum^T_{t=0} r(s_t, a_t) ] \\
    &\text{s.t. } \E_{s_t, a_t \sim \pi_\phi} [-\log{\pi_t(a_t|s_t)}] \ge \mathbb{H}_U, \forall t
\end{split}
\end{equation}
where the sum of rewards are maximized subject to the entropy constraint, with the minimum target expected entropy $\mathbb{H}_U$ fixed as the entropy of a uniform policy across all discrete actions. By setting the following objective with a loose upper limit on the target entropy, $\alpha$ can be obtained by minimizing:
\begin{equation}
    J(\alpha) = \E_{(a|s) \sim \pi} [\alpha(-\log( \pi_\phi(a_t|s_t) ) - \mathbb{H}_U ]
\label{alpha_update}
\end{equation}

\subsection{Stabilizing Discrete Soft-Actor Critic}
Previous studies have shown that vanilla DSAC is not immune to problems such as policy instability and Q-value underestimation. For e.g., the temperature parameter $\alpha$ is directly tied to the agent's ability to explore and a poor initialization of the target entropy can hurt the automated value-entropy tuning and overall performance \cite{wang2020meta, Xu2021TargetEA}. \cite{Zhou2022RevisitingDS} have proposed counter-measures such as introducing a entropy-penalty term and double average clipped Q-learning. In our work, we found the double average Q-learning trick to be useful in mitigating state value underestimation. Specifically, for the learning of the critic, the soft value estimates become:
\begin{equation}
     V(s_t) = \E_{a_t \sim \pi} [ \overline{Q}_{\theta_i(s_{t+1}, a_{t+1})} - \alpha \log(\pi_\phi(s_t)) ]
\end{equation}
where the $\min$ operator in \cref{soft_value_estimate} is replaced by the average operator $\overline{Q}_{\theta_i(s_{t+1}, a_{t+1})}$ in the double-critic trick.

\section{DSAC-C: Constrained Discrete Soft-Actor Critic} 
Following \cref{max_ent_method}, the unconstrained MaxEnt RL objective only considers the entropy term. In this section, we propose additional constraints to the SAC objective function, with the target expected value constraints estimated from a surrogate critic policy. 

\subsection{Boltzmann Critic Policy}
From the SAC objective, it can be shown that the greedy soft policy is simply a $\softmax$ over it's Q values scaled by the temperature parameter $\alpha$. As the temperature $\alpha \rightarrow 0$, the softmax operator simply becomes the $\max$ operator. Following this, we propose to introduce a surrogate Boltzmann policy $\pi_\theta$ derived directly from the Q-estimates. We highlight that this is a separate surrogate policy from the critic and it is not to be confused with the actor's policy $\pi_\phi$. This surrogate critic policy serves as a supplementary form of guidance to the actor during training.

\begin{equation}
    \pi_\theta(a_{t}|s_{t}) = \frac{  e^{ Q_\theta(s_{t}, a_{t})/\alpha } }{ \sum e^{ Q_\theta(s_{t}, a_{t})/\alpha }}
\label{actor_policy}
\end{equation}
\proof{See \cref{SAC}}

\subsection{Mean Constrained Soft-Actor Critic}
Recall that the MaxEnt method in \cref{max_ent_method}, allows for constraints to be placed to the maximum entropy objective. Using the surrogate critic policy, we propose the constraints to be the expected mean value targets for each state-action pair $s_t, a_t$ computed batch wise from the critics as $\mu_\theta = \E_{Q_\theta}[Q_\theta(s_{t}, a_{t}) ]$. The computed constraints are appended to \cref{cont_actor_update} with the newly proposed update for the mean constrained soft actor given by the following objective function:

\begin{equation}
    \begin{split}
    &J_\pi(\phi) = \E_{s_t \sim D} [[\alpha \log( \pi_\phi(a_t|s_t) ) - Q_\theta(s_t, a_t) ] \\
    & \qquad \qquad + \lambda_1 ( \underbrace{ \E_{Q_\phi} [Q_\theta(s_{t}, a_{t})] - \mu_\theta }_{\text{Mean constraint}})]
\end{split}
\label{mean_actor_update}
\end{equation}
where the predicted probabilities of the actor and Q-values are used to compute it's expected mean given as $\E_{Q_\phi} [Q_\theta(s_{t}, a_{t})]$ which is minimized against the critic's mean. $\lambda_1$ is the respective Lagrange multiplier for the mean constraint which can be computed numerically e.g. using Newton Raphson method for each given state sample per minibatch. Consequently, using the objective function in \cref{mean_actor_update}, the new policy for the mean constrained actor is given as:
\begin{equation}
    \pi^\mu_\phi(a_{t}|s_{t}) = \frac{  e^{ \bigl( Q_\theta(s_{t}, a_{t}) - \lambda_1 Q_\theta(s_{t}, a_{t}) \bigr)/\alpha } }{ \sum e^{ \bigl( Q_\theta(s_{t}, a_{t}) - \lambda_1 Q_\theta(s_{t}, a_{t}) \bigr) /\alpha }}
\label{actor_policy_mean}
\end{equation}
\proof{See \cref{Mean_SAC}}

\begin{algorithm*}[!t]
\SetAlgoNoLine
\SetAlgoNoEnd
\DontPrintSemicolon

\KwData{Initialize memory replay buffer $D$;}
\BlankLine
Initialize learning rates, $\beta_Q, \beta_\pi$, network parameters $\theta_1, \theta_2, \phi$ and univariate Gaussian $\mathcal{N}$;

\SetKwFunction{fOptimize}{Optimize}
\SetKwFunction{fLabelFusion}{LabelFusion}
\SetKwFunction{fNewtonRaphson}{NewtonRaphson}

\lFor{each iteration}{\;

\hskip2.0em \lFor{each environment timestep}{\;
\hskip4.0em $a_t \sim \pi_\phi(a_t|s_t)$ \tcp*{Sample action following actor's policy}
\hskip4.0em $s_{t+1} \sim p(s_{t+1} s_t, a_t)$ \tcp*{Sample transition from the environment}
\hskip4.0em $D \leftarrow D \cup \{( s_t, a_t, r(s_t, a_t), s_{t+1} ) \} $ \tcp*{Store transition into memory buffer}
}
\hskip2.0em \lFor{each gradient update step}{\;
\hskip4.0em $\theta_i \leftarrow \theta_i - \beta_Q \hat{\Delta}_{\theta_i} J(\theta_i) - \text{for $i$} \in \{1,2\} $ \tcp*{Update the critic's parameters}
\hskip4.0em $\epsilon \leftarrow \mathcal{N}(0, 1) $ \tcp*{Randomly sample $\epsilon$ from $\mathcal{N}$ and add to constraints}
\hskip4.0em $\phi \leftarrow \phi - \beta_\pi \hat{\Delta}_{\phi} J_\pi(\Phi)  $ \tcp*{Update the actor's parameters}
\hskip4.0em $\lambda_i \leftarrow \lambda_i +  \Delta \lambda_i $\tcp*{Solve $\lambda_i$ numerically using $\fNewtonRaphson{}$}
\hskip4.0em $\alpha \leftarrow \alpha - \beta_\pi \hat{\Delta}_{\alpha} J_\alpha(\alpha) $ \tcp*{Adjust temperature}
\hskip4.0em $\bar{Q_i} \leftarrow \tau Q_i + (1 - \tau) \bar{Q_i} - \text{for $i$} \in \{1,2\} $  \tcp*{Update target critics parameters}
}
\Return{$\theta_1, \theta_2, \phi $} 
}
\BlankLine

\Func{\fNewtonRaphson{}}{
$\delta$ = 1e-15 \;
\While{ g($\lambda$) $> \delta$} {
$\lambda_{n+1} = \lambda_n -\frac{g(\lambda)}{g'(\lambda)} $ \tcp*{Update Lagrange multipliers $\lambda_i$} }
\Return $\lambda_i$
}
\caption{The DSAC-C Algorithm}
\label{algo:algorithm}
\end{algorithm*}

\subsection{Variance Constrained Soft-Actor Critic}
Similar to the steps proposed in the expected mean constraint, the expected variance targets can be computed from the critics as $\sigma^2_\theta = \E_{Q_\theta}[ (Q_\theta(s_{t}, a_{t}) - \mu )^2 ]$. Using this definition, we can reuse the expected mean targets computed earlier, using the handy relationship $\sigma^2_\theta = \E_{Q_\theta}[ Q_\theta(s_{t}, a_{t})^2 ] - \E_{Q_\theta}[  Q_\theta(s_{t}, a_{t}) ]^2$. Finally, the new objective function for the variance constrained actor is given as:
\begin{equation}
    \begin{split}
    & J_\pi(\phi) = \E_{s_t \sim D} [[\alpha\log( \pi_\phi(s_t)) - Q_\theta(s_t, a_t)] \\
    & \qquad \qquad + \lambda_2 (\underbrace{ \E_{Q_\phi} [(Q_\theta(s_t, a_t) - \mu_\phi)^2] - \sigma^2_\theta}_{\text{Variance constraint}}) ]
    \end{split}
    \label{variance_actor_update}
\end{equation}
$\lambda_2$ is the corresponding Lagrange multiplier for the variance constraint, with the new policy for the variance constraint actor given as:
\begin{equation}
    \pi^{\sigma^2}_\phi(a_{t}|s_{t}) = \frac{  e^{ \bigl(Q_\theta(s_{t}, a_{t}) - \lambda_2 (Q_\theta(s_{t}, a_{t}) - \mu)^2 \bigr)/\alpha } }{ \sum e^{ \bigl( Q_\theta(s_{t}, a_{t}) - \lambda_2 (Q_\theta(s_{t}, a_{t}) - \mu)^2 \bigr) /\alpha }}
\label{actor_policy_variance}
\end{equation}
\proof{See \cref{Variance_SAC}}

We highlight that our proposed methods are not equivalent to naively minimizing the cross-entropy between the actor's policy and surrogate critic's policy. Since the KL-divergence between the actor and critic is already considered in the original SAC objective, the additional constraints serve as an additional form of guidance from the critic during training. Combining all of the above proposed changes, we summarize our method in Algorithm \ref{algo:algorithm} and show the respective mathematical derivations in the Appendix.

\subsection{Perturbed Expected Value Targets}
For RL tasks, introducing parametric noise during training has been shown to improve agent performance and encourage exploration \cite{Fortunato2017NoisyNF, Plappert2017ParameterSN}. In this work, we follow a similar chain-of-thought by perturbing Q-value estimates of the critics before computing the characteristic constraints. By randomly sampling a small noise $\epsilon$ from a univariate Gaussian $\mathcal{N}$, we inject $\epsilon$ into the Q-values before computing the expected value targets. Specifically, the mean constrainted actor in \cref{mean_actor_update} becomes:
\begin{equation}
    \begin{split}
    &J_\pi(\phi) = \E_{s_t \sim D} [[\alpha \log( \pi_\phi(a_t|s_t) ) - Q_\theta(s_t, a_t) ] \\
    & \qquad \qquad + \lambda_1 ( \underbrace{ \E_{Q_\phi} [Q_\theta(s_{t}, a_{t}) + \epsilon] - \mu_\theta }_{\text{Perturbed mean constraint}})]
\end{split}
\end{equation}
and the variance constrained actor in \cref{variance_actor_update} becomes:
\begin{equation}
    \begin{split}
    & J_\pi(\phi) = \E_{s_t \sim D} [[\alpha\log( \pi_\phi(s_t)) - Q_\theta(s_t, a_t)] \\
    & \qquad \qquad + \lambda_2 (\underbrace{ \E_{Q_\phi} [(Q_\theta(s_t, a_t) - \mu_\phi + \epsilon)^2] - \sigma^2_\theta}_{\text{Perturbed variance constraint}}) ]
    \end{split}
\end{equation}
\subsection{Taming of the Q}
Q-value estimates tend to grow larger as the agent becomes better at solving tasks. Which can be problematic for the newly proposed policies, since computing the expectation involves taking the exponent of large Q-values resulting in problems such as overflow and policy instability. To alleviate these issues, we propose the following workarounds to better manage large Q-values.

\begin{table*}[!htb]
\small
\center
\begin{adjustbox}{width=\textwidth}
\begin{tabular}{cc|ccccccccc|}
&Environment &Alien &Amidar & BattleZone& BeamRider &Breakout &CrazyClimber &MsPacman &Seaquest &UpNDown\\
\hline
\multirow{3}{*}{\rotatebox[origin=c]{90}{ID}}
&DSAC &183.00 $\pm21$ &12.20 $\pm1$ &2975.53 $\pm742$ &806.80 $\pm324$ &27.51 $\pm3$ &434.53 $\pm445$ &403.00 $\pm47$ &232.90 $\pm262$ &805.42 $\pm84$\\ 
&DSAC-M (Ours) &\textbf{232.0} $\pm13$ &\textbf{15.30} $\pm8$ &\textbf{4234.07} $\pm840$ &\textbf{1749.00} $\pm266$ &\textbf{47.07} $\pm5$ &\textbf{605.43} $\pm380$ &\textbf{420.67} $\pm63$ &\textbf{281.68} $\pm259$ &\textbf{2521.67} $\pm1411$\\ 
&DSAC-V (Ours) &215.8 $\pm41$ &4.35 $\pm5$ &2233.83 $\pm602$ &417.48 $\pm174$ &29.03 $\pm2$ &155.27 $\pm132$ &357.42 $\pm83$ &137.20 $\pm79$ &1079.08 $\pm288$\\
\hline
\multirow{3}{*}{\rotatebox[origin=c]{90}{Snow}} 
&DSAC &166.17 $\pm27$ &11.15 $\pm1$ &1750.71 $\pm648$ &650.23 $\pm246$ &8.03 $\pm1$ &378.60 $\pm401$ &359.50 $\pm54$ &225.81 $\pm254$ &731.92 $\pm111$\\
&DSAC-M (Ours) &\textbf{221.08} $\pm37$ &\textbf{13.04} $\pm8$ &\textbf{3641.88} $\pm607$ &\textbf{1522.82} $\pm304$ &\textbf{8.83} $\pm1$ &\textbf{705.11} $\pm453$ &\textbf{394.33} $\pm39$ &\textbf{282.14} $\pm253$ &\textbf{2249.25} $\pm1092$ \\
&DSAC-V (Ours) &206.83 $\pm42$ &4.27 $\pm5$ &1559.02 $\pm761$ &363.90 $\pm130$ &8.16 $\pm1$ &80.09 $\pm66$ &332.58 $\pm74$ &141.26 $\pm69$ &970.83 $\pm179$\\
\hline
\multirow{3}{*}{\rotatebox[origin=c]{90}{Rain}}
&DSAC &170.00 $\pm12$ &12.27 $\pm1$ &2725.26 $\pm247$ &659.30 $\pm255$ &22.97 $\pm1$ &377.69 $\pm401$ &354.00 $\pm57$ &231.88 $\pm257$ &794.83 $\pm84$\\
&DSAC-M (Ours) &\textbf{217.67} $\pm23$ &\textbf{13.43} $\pm8$ &\textbf{4425.32} $\pm270$ &\textbf{1532.36} $\pm249$ &\textbf{33.95} $\pm2$ &\textbf{651.29} $\pm413$ &\textbf{392.83} $\pm32$ &\textbf{277.70} $\pm257$ &\textbf{2404.92}$\pm1237$\\
&DSAC-V (Ours) &203.42 $\pm50$ &4.93 $\pm5$ &1692.00 $\pm768$ &346.97 $\pm152$ &24.72 $\pm2$ &94.22 $\pm73$ &345.33 $\pm70$ &135.29 $\pm71$ &1008.83 $\pm247$\\
\hline
\multirow{3}{*}{\rotatebox[origin=c]{90}{Fog}}
&DSAC &166.25 $\pm12$ &\textbf{12.67} $\pm1$ &2141.80 $\pm574$ &543.93 $\pm244$ &6.99 $\pm2$ &312.62 $\pm294$ &371.00 $\pm23$ &229.61 $\pm251$ &776.58 $\pm71$\\
&DSAC-M (Ours) &\textbf{224.75} $\pm45$ &10.28 $\pm6$ &\textbf{2958.57} $\pm510$ &\textbf{1249.37} $\pm281$ &\textbf{7.00} $\pm0$ &\textbf{321.68} $\pm210$ &\textbf{374.92} $\pm58$ &\textbf{280.16} $\pm256$ &\textbf{4234.07} $\pm840$\\
&DSAC-V (Ours) &208.58 $\pm50$ &4.84 $\pm4$ &2341.95 $\pm802$ &241.43 $\pm88$ &5.99 $\pm0$ &75.03 $\pm61$ &341.25 $\pm57$ &138.33 $\pm82$ &2233.83 $\pm602$\\
\hline 
&Overall &\textbf{DSAC-M} &\textbf{DSAC-M} &\textbf{DSAC-M} &\textbf{DSAC-M} &\textbf{DSAC-M} &\textbf{DSAC-M} &\textbf{DSAC-M} &\textbf{DSAC-M} &\textbf{DSAC-M}\\
\hline
\end{tabular}
\end{adjustbox}
\caption{Raw return scores at 1,000,000 steps for each method benchmarked on Atari games across three random seeds. The best median scores are in bold with $\pm$ indicating 1 standard deviation. DSAC-M achieves the best performance on nine out of nine environments across all distribution shifts. }
\label{main_table}
\end{table*}

\begin{figure}[!tb]
    \small
    \centering
    \includegraphics[width=\columnwidth]{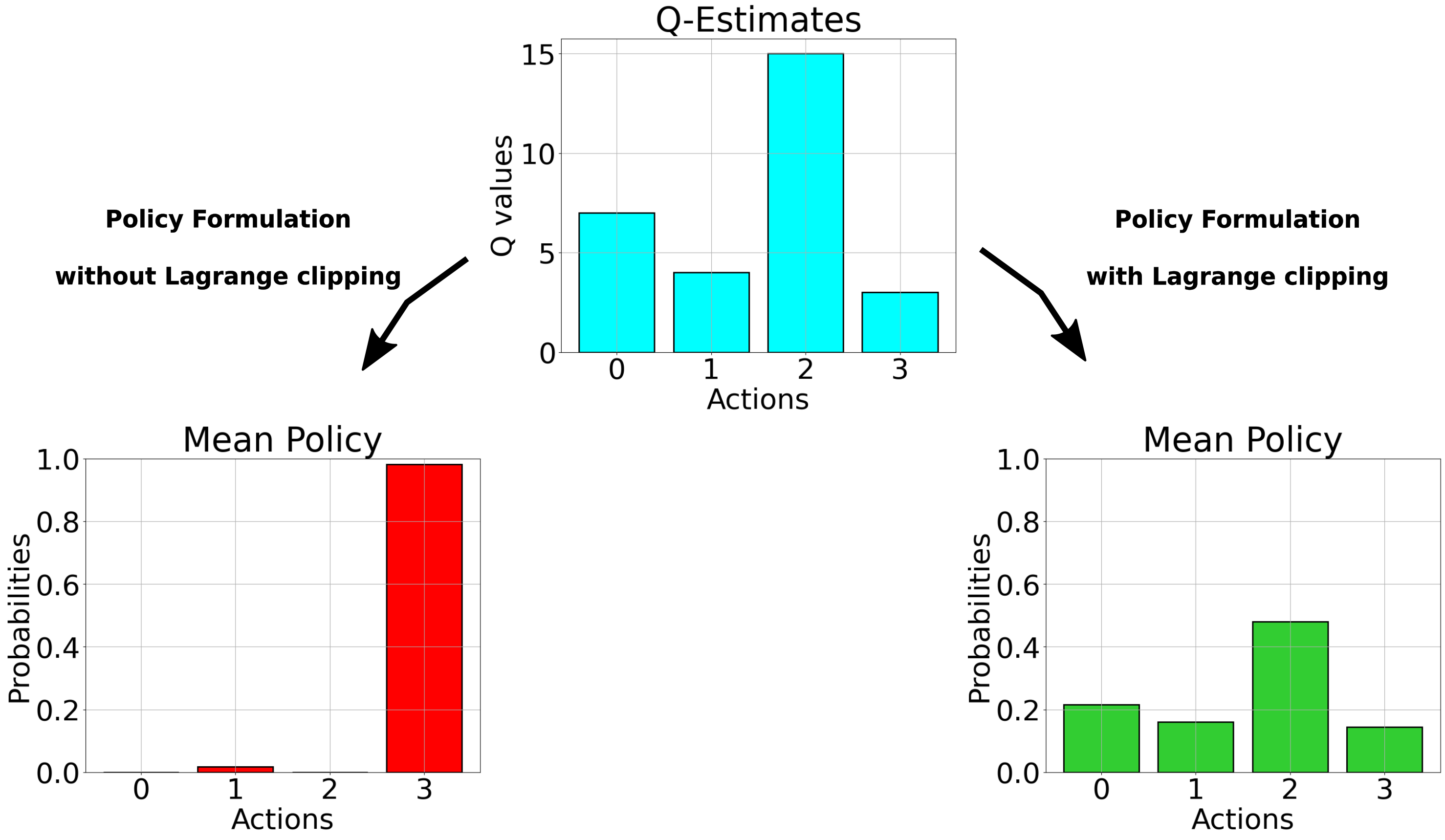}
    \caption{With Lagrange multiplier clipping, the constrained agent's policy is stable and aligned to the Q-values (green). Without clipping, the policy/argmax is lost (red).}
\label{lagrange clip}
\end{figure}
\noindent \textbf{Lagrange Multiplier Clip:} During the numerical finding of $\lambda_i$, these estimates can grow too large or turn negative, causing a drastic change/flipping of policies. \cref{lagrange clip} shows an example failure case of the mean policy when $\lambda_1$ is too large, causing a change in the argmax of the greedy policy. Therefore, we propose to clip $\lambda_i \in [0-1]$, which bounds the constrained policy within small positive values. Specifically, if $\lambda_1 = 0$ the original SAC policy in \cref{actor_policy} is obtained and the maximum penalty for the constraints are applied when $\lambda_1 = 1$.

\noindent \textbf{Stable Softmax Trick:} To avoid overflow with large Q-values, we also apply the stable softmax trick. This involves the subtraction of the maximum element in the Q-values before raising the exponent. As an example, the surrogate critic policy in \cref{actor_policy} becomes $\frac{  e^{ Q_\theta(s_{t}, a_{t}) - \max(Q_\theta(s_{t}, a_{t})) /\alpha } }{ \sum e^{ Q_\theta(s_{t}, a_{t}) - \max(Q_\theta(s_{t}, a_{t}))/\alpha }}$. This trick ensures that the resulting exponent is mathematically equivalent to the original policy and does not affect gradient computation. We apply this trick to Equations (\ref{actor_policy},\ref{actor_policy_mean},\ref{actor_policy_variance}). 

\section{Experiments and Results}
In this section, we demonstrate the overall performance gains from the proposed constrained variants of DSAC on both ID and augmented OOD Atari. We further discuss the effects of these constraints on the policy's entropy and expected value errors.

\subsection{Experimental Setup}
We benchmark our proposed algorithm against agents trained with vanilla D-SAC \cite{Christodoulou2019SoftAF} on nine different games from the Atari 2600 arcade learning environment \cite{Bellemare2012TheAL}. We begin each game with up to 30 no-op actions, so as to allow the agent a random starting position as per \cite{mnih2013playing}. For our evaluation we follow the protocols of the respective author and test the deterministic policy of the agent after every 1M steps (4M frames) for 10 episodes with a cap of 5mins or 18000 frames across three different random seeds. Our experiments are conducted on a NVIDIA RTX A5000 with Intel Xeon Silver 4310 CPU.

For our experiments we use CleanRL \cite{huang2022cleanrl}, with additional details such as hyper-parameters included in the Appendix. We use imgaug \cite{imgaug} to create the OOD shifted form of Atari, the augmentations are only applied during test time. We highlight that other forms of image augmentations (i.e. Gaussian Noise, Contrast, Saturation) can be used to evaluate agents, however we chose "Snow", "Rain" and "Fog" to simulate the more likely shifts in the real world. 

\begin{figure*}[!tb]
    \centering
    \includegraphics[width=\textwidth]{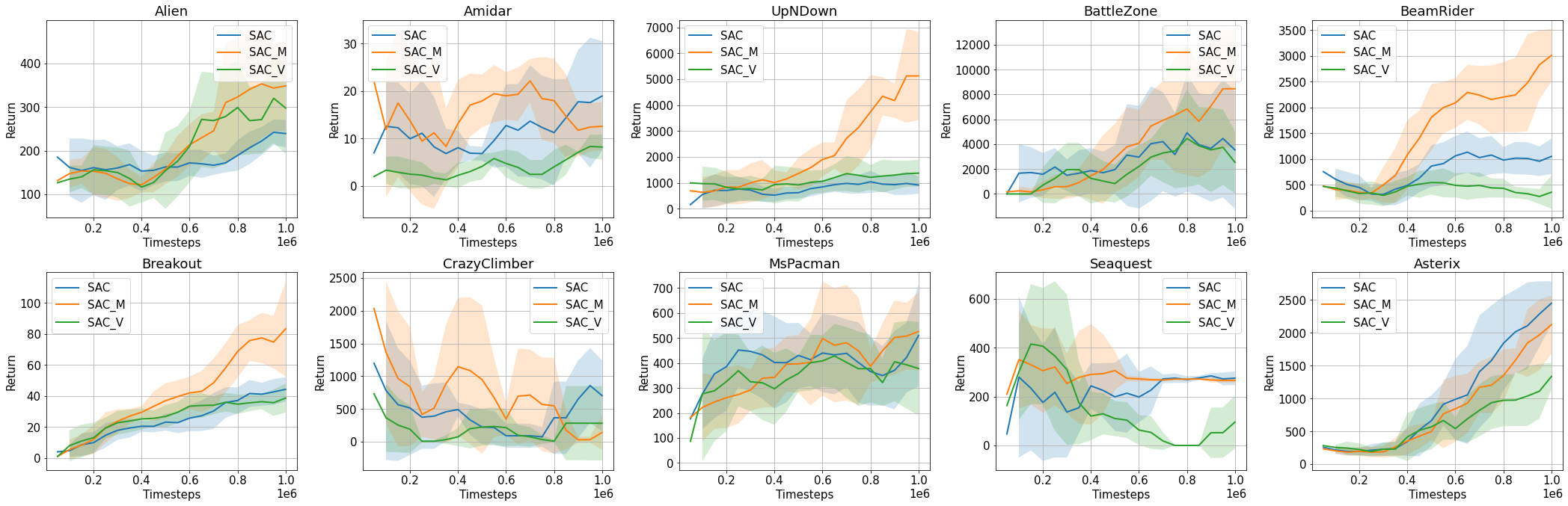}
    \caption{The learning curves on ten different Atari benchmarks, showing how additional constraints can be used to improve the overall performance of the SAC algorithm. Plots are smoothed for readability by a moving average of factor 10.}
\label{Atari results}
\end{figure*}

\subsection{ID Atari Evaluation}
We compare our proposed mean and variance constraint forms against the baseline DSAC. We show the learning curves for each environment along with the standard deviation across seeds indicated by shaded confidence bounds in \cref{Atari results}. From our experiments, we find that the variance constraint form improves agent learning on three ID environments, whereas the mean constraint form is able to improve agent performance on nine out of the nine ID environments evaluated. We report the highest improvements in relative performance coming from the mean constraint form on solving environments such as ``UpNDown", ``Breakout" and ``BattleZone" by about 213\%, 71\% and 42\% respectively.

\begin{figure*}[!htb]
  \begin{subfigure}{0.24\textwidth}
    \includegraphics[width=\linewidth]{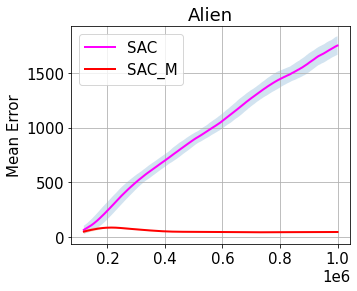}
    \caption{Expected Mean Error} \label{fig:1a}
  \end{subfigure}%
  \hspace*{\fill}   
  \begin{subfigure}{0.24\textwidth}
    \includegraphics[width=\linewidth]{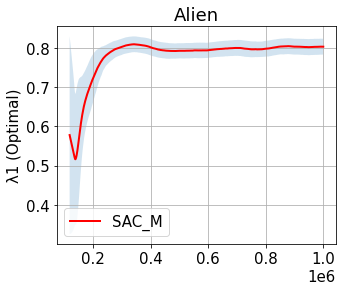}
    \caption{Optimal $\lambda_1$} \label{fig:1b}
  \end{subfigure}%
  \begin{subfigure}{0.24\textwidth}
    \includegraphics[width=\linewidth]{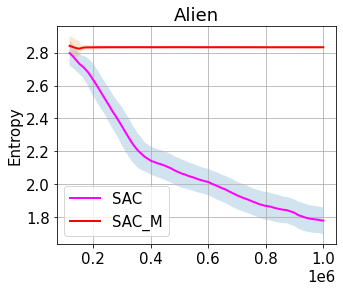}
    \caption{Entropy of Policy} \label{fig:1c}
  \end{subfigure}%
  \hspace*{\fill}  
  \begin{subfigure}{0.24\textwidth}
    \includegraphics[width=\linewidth]{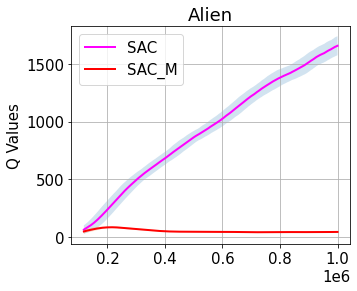}
    \caption{Q-estimates} \label{fig:1d}
  \end{subfigure}%
\caption{Additional constraints between the actor and critic's expected values affects the policy's entropy and Q-estimates during training.} \label{EVT_Entropy}
\end{figure*}

\subsection{OOD Atari Evaluation}
Apart from evaluating our agent on ID Atari, we further demonstrate the performance gains of additional constraints on the makeshift OOD form of Atari. Similar to the ID environments, we also observe improvements with the mean constraint form for nine out of the nine environments for each of the three domain shifts respectively. 

For the final overall performance, we equally consider the scores of each agent by summing their ID and OOD scores in the last row of \cref{main_table}. We find that with additional constraints, we are able to improve the general performance of DSAC for all nine games on both ID and OOD scenarios. 

\subsection{Actor Entropy and Expected Value errors}
We further examine the effects of the added mean constraint, by plotting the expected mean errors along with the policy's entropy, optimal values for $\lambda_1$ and Q-estimates in \cref{EVT_Entropy}. Specifically, we measure the expected errors by taking the L1 difference between the actor and critic's expected mean values. For these plots, we evaluate the agents on the ID environment of ``Alien" .

In \cref{fig:1a} and \cref{fig:1b}, our findings demonstrate that the inclusion of constraints lead to lower expected value errors between the actor and critic with the optimal value for $\lambda_1$ converging to about 0.8. These constraints help the actor better learn from the critic during training. \cref{fig:1c} shows that the added constraints lead to higher policy entropy, which in turn helps improve agent exploration and robustness OOD. However, this can negatively cause the critic to exhibit lower Q-value estimates, which 
might be unhelpful in situations where Q-values are needed. Regardless, these lower Q-estimates do not affect the policy/actor and can be utilized in actor-critic methods.

\section{Conclusion}
We revisited a classical idea from the Principle of Maximum Entropy and proposed a novel addition to the SAC family using constraints derived from the critic. Our method includes modifications made to the actor's objective function and automatic solving of the Lagrange multipliers for the constraints. We also provide theoretical support for the new policy objectives and propose practical strategies to stabilize our method. Next, we benchmarked the overall performance gains against discrete SAC on both ID and OOD Atari. Finally, we discussed the different effects of the constraints on the expected value errors, actor's entropy and Q-value estimates.

\section{Acknowledgements}
This research was supported in part by the Ministry of Education, Singapore, under the MOE AcRF TIER 3 Grant (MOE-MOET32022-0001), and by the DSO-AISG Incentive programme. The authors also extend their gratitude to Tan Yingkiat and Bai Yunwei for their valuable suggestions and feedback.

\bibliographystyle{IEEEtran}
\bibliography{IEEEabrv, ijcnn}

\clearpage
\medskip
\onecolumn

\appendix
\small 
\section{Appendix}
\subsection{Discrete Soft-Actor Critic}
\label{SAC}
\begin{align*}
   &J_\pi(\phi) = \sum \alpha \pi_\phi(s_t)\log \pi_\phi(s_t) - \pi_\phi(s_t)Q_\theta(s_t, a_t) + \lambda_0 ( \sum \pi_\phi(a_t|s_t) - 1 ) \\
   &\textit{Expand the Lagrangian, take its derivative and set to zero}\\
   &\frac{\partial J_\pi(\phi)}{\partial \pi_\phi(s_t)} = \alpha [\log \pi_\phi(s_t) + 1] - Q_\theta(s_t, a_t) + \lambda_0 = 0 \tag{$1$}\\
   &\textit{Subject to the following constraints}\\
   &\sum \pi_\phi(a_t|s_t) = 1 \tag{$2$}\\
   &\textit{Rearrange Equation 1 and obtain general solution}\\
   &\pi_\phi(s_t) = e^{- 1 - \lambda_0 + Q_\theta(s_t, a_t)/\alpha } \tag{$3$}\\
   &\textit{Substitute general solution into first constraint }\\
   &\pi_\phi(s_t) = \frac{1}{Z} \sum e^{Q_\theta(s_t, a_t)/\alpha } = 1 \tag{Let $Z = e^{1 + \lambda_0}$ and substitute \underline{3} into \underline{2}} \\
   &\textit{Rearrange and obtain the soft actor's policy}\\
   &\pi_\phi(s_t) = \frac{e^{Q_\theta(s_t, a_t)/\alpha }}{\sum e^{Q_\theta(s_t, a_t)/\alpha }} \tag{$4$}\\ 
\end{align*}

\subsection{Mean Constraint form}
\label{Mean_SAC}
\begin{align*}
   &J_\pi(\phi) = \sum \pi_\phi(s_t)\log \pi_\phi(s_t) - \pi_\phi(s_t)Q_\theta(s_t, a_t) + \lambda_0 ( \sum \pi_\phi(a_t|s_t) - 1 ) + \lambda_1 ( \sum \pi_\phi(a_t|s_t) Q_\theta(s_{t}, a_{t}) - \mu_\theta )\\
   &\textit{Expand the Lagrangian, take its derivative and set to zero}\\
   &\frac{\partial J_\pi(\phi)}{\partial \pi_\phi(s_t)} = \log \pi_\phi(s_t) + 1 - Q_\theta(s_t, a_t) + \lambda_0 + \lambda_1 Q_\theta(s_t, a_t) = 0 \tag{$1$}\\
   &\textit{Subject to the following constraints}\\
   &\sum \pi_\phi(a_t|s_t) = 1 \tag{$2$}\\
   &\sum \pi_\phi(a_t|s_t) Q_\theta(s_{t}, a_{t}) = \mu_\theta \tag{$3$}\\
   &\textit{Rearrange Equation 1 and obtain general solution}\\
   &\pi_\phi(s_t) = e^{- 1 - \lambda_0 + Q_\theta(s_t, a_t) - \lambda_1 Q_\theta(s_t, a_t) } \tag{$4$}\\
   &\textit{Substitute general solution into first constraint }\\
   &\pi_\phi(s_t) = \frac{1}{Z} \sum e^{Q_\theta(s_t, a_t) - \lambda_1 Q_\theta(s_t, a_t) } = 1 \tag{Let $Z = e^{1 + \lambda_0}$ and substitute \underline{4} into \underline{2}} \\
   &\textit{Rearrange and obtain the actor's policy}\\
   &\pi_\phi(s_t) = \frac{e^{Q_\theta(s_t, a_t) - \lambda_1 Q_\theta(s_t, a_t)}}{\sum e^{Q_\theta(s_t, a_t) - \lambda_1 Q_\theta(s_t, a_t) }} \tag{$5$}\\ 
   &\textit{Satisfy second constraint}\\
   &\sum \pi_\phi(s_t) Q_\theta(s_t, a_t) = \frac{e^{Q_\theta(s_t, a_t) - \lambda_1 Q_\theta(s_t, a_t)}}{\sum e^{Q_\theta(s_t, a_t) - \lambda_1 Q_\theta(s_t, a_t) }} Q_\theta(s_t, a_t) = \mu_\theta \tag{Substitute policy into \underline{3}}\\
   &\textit{Solve for $\lambda_1$ using Newton's method} \\
   &g(\lambda_1) = \frac{e^{Q_\theta(s_t, a_t) - \lambda_1 Q_\theta(s_t, a_t)}}{\sum e^{Q_\theta(s_t, a_t) - \lambda_1 Q_\theta(s_t, a_t) }} Q_\theta(s_t, a_t) - \mu_\theta \\
   &\textit{Take derivative of $g(\lambda_1)$ wrt $\lambda_1$ using the quotient rule} \\
   &g'(\lambda_1) = \frac{-Q_\theta(s_t, a_t)^2 e^{Q_\theta(s_t, a_t) - \lambda_1 Q_\theta(s_t, a_t)} \sum e^{Q_\theta(s_t, a_t) - \lambda_1 Q_\theta(s_t, a_t) } + \bigl( Q_\theta(s_t, a_t)^2 e^{Q_\theta(s_t, a_t) - \lambda_1 Q_\theta(s_t, a_t)} \bigr)^2 }{{ \bigl( \sum e^{Q_\theta(s_t, a_t) - \lambda_1 Q_\theta(s_t, a_t) }}  \bigr) ^2} \\
   &\textit{Rearrange and obtain the final step for the derivative} \\
   &g'(\lambda_1) = -\sum Q_\theta(s_t, a_t)^2 \frac{e^{Q_\theta(s_t, a_t) - \lambda_1 Q_\theta(s_t, a_t)}}{\sum e^{Q_\theta(s_t, a_t) - \lambda_1 Q_\theta(s_t, a_t) }} + \bigl( \sum Q_\theta(s_t, a_t) \frac{e^{Q_\theta(s_t, a_t) - \lambda_1 Q_\theta(s_t, a_t)}}{\sum e^{Q_\theta(s_t, a_t) - \lambda_1 Q_\theta(s_t, a_t) }}  \bigr) ^2 \\
   &g'(\lambda_1) = -Var[Q_\theta] \tag{Which gives the -ve Variance of the estimated Q-values}\\
\end{align*}
\clearpage

\subsection{Variance Constraint form}
\label{Variance_SAC}
\begin{align*}
   &J_\pi(\phi) = \sum \pi_\phi(s_t)\log \pi_\phi(s_t) - \pi_\phi(s_t)Q_\theta(s_t, a_t) + \lambda_0 ( \sum \pi_\phi(a_t|s_t) - 1 ) + \lambda_2 (\sum \pi_\phi(a_t|s_t) [Q_\theta(s_t, a_t) - \mu_\phi]^2 - \sigma^2_\theta )\\
   &\textit{Expand the Lagrangian, take its derivative and set to zero}\\
   &\frac{\partial J_\pi(\phi)}{\partial \pi_\phi(s_t)} = \log \pi_\phi(s_t) + 1 - Q_\theta(s_t, a_t) + \lambda_0 + \lambda_2 [Q_\theta(s_t, a_t) - \mu_\phi]^2 = 0 \tag{$1$}\\
   &\textit{Subject to the following constraints}\\
   &\sum \pi_\phi(a_t|s_t) = 1 \tag{$2$}\\
   &\sum \pi_\phi(a_t|s_t) [Q_\theta(s_t, a_t) - \mu_\phi]^2 = \sigma^2_\theta \tag{$3$}\\
   &\textit{Rearrange Equation 1 and obtain general solution}\\
   &\pi_\phi(s_t) = e^{- 1 - \lambda_0 + Q_\theta(s_t, a_t) - \lambda_2 [Q_\theta(s_t, a_t) - \mu_\phi]^2 } \tag{$4$}\\
   &\textit{Substitute general solution into first constraint }\\
   &\pi_\phi(s_t) = \frac{1}{Z} \sum e^{Q_\theta(s_t, a_t) - \lambda_2 [Q_\theta(s_t, a_t) - \mu_\phi]^2 } = 1 \tag{Let $Z = e^{1 + \lambda_0}$ and substitute \underline{4} into \underline{2}} \\
   &\textit{Rearrange and obtain the actor's policy}\\
   &\pi_\phi(s_t) = \frac{e^{Q_\theta(s_t, a_t) - \lambda_2 [Q_\theta(s_t, a_t) - \mu_\phi]^2 }}{\sum e^{Q_\theta(s_t, a_t) - \lambda_2 [Q_\theta(s_t, a_t) - \mu_\phi]^2 }} \tag{$5$}\\
   &\textit{Satisfy second constraint}\\
   &\sum \pi_\phi(s_t) [Q_\theta(s_t, a_t) - \mu_\phi]^2 = \frac{e^{Q_\theta(s_t, a_t) - \lambda_2 [Q_\theta(s_t, a_t) - \mu_\phi]^2 }}{\sum e^{Q_\theta(s_t, a_t) - \lambda_2 [Q_\theta(s_t, a_t) - \mu_\phi]^2 }} [Q_\theta(s_t, a_t) - \mu_\phi]^2 = \sigma^2_\theta \tag{Substitute policy into \underline{3}}\\
   &\textit{Solve for $\lambda_2$ using Newton's method} \\
   &g(\lambda_2) = \frac{e^{Q_\theta(s_t, a_t) - \lambda_2 [Q_\theta(s_t, a_t) - \mu_\phi]^2 }}{\sum e^{Q_\theta(s_t, a_t) - \lambda_2 [Q_\theta(s_t, a_t) - \mu_\phi]^2 }} [Q_\theta(s_t, a_t) - \mu_\phi]^2 - \sigma^2_\theta  \\
   &\textit{Take derivative of $g(\lambda_2)$ wrt $\lambda_2$ using the quotient rule} \\
   &g'(\lambda_2) = \frac{ -[Q_\theta - \mu_\phi]^4 e^{Q_\theta - \lambda_2[Q_\theta - \mu_\phi]^2 } \sum e^{Q_\theta - \lambda_2 [Q_\theta- \mu_\phi]^2 } + \bigl( [Q_\theta - \mu_\phi] ^2 e^{Q_\theta - \lambda_2 [Q_\theta - \mu_\phi]^2 } \bigr)^2 }{{ \bigl( \sum e^{Q_\theta - \lambda_2 [Q_\theta - \mu_\phi]^2 }}  \bigr) ^2} \\
   &\textit{Rearrange and obtain the final step for the derivative} \\
   &g'(\lambda_2) = -\sum [Q_\theta - \mu_\phi]^4 \frac{e^{Q_\theta - \lambda_2 [Q_\theta - \mu_\phi] }}{\sum e^{Q_\theta - \lambda_2 [Q_\theta - \mu_\phi] }} + \bigl( \sum [Q_\theta - \mu_\phi]^2 \frac{e^{Q_\theta - \lambda_2 [Q_\theta - \mu_\phi]^2 }}{\sum e^{Q_\theta - \lambda_2 [Q_\theta - \mu_\phi]^2 }}  \bigr) ^2 \\
   &g'(\lambda_2) = -Kur[Q_\theta] \tag{Which gives the -ve Kurtosis of the estimated Q-values}\\
\end{align*}

\end{document}